%% file: main_arxiv.tex
\documentclass[10pt,twocolumn,letterpaper]{article}

\usepackage{iccv}              

\input{macros}

\usepackage{graphicx}
\usepackage{amsfonts,amssymb,amsmath,amsthm}
\usepackage{booktabs}
\usepackage{xcolor}
\usepackage{color}
\usepackage{acronym} 
\usepackage{float}
\usepackage{multirow}
\usepackage{bm}

\usepackage{bbding}
\usepackage{pifont}
\usepackage{wasysym}
\usepackage{enumitem}
\usepackage{tabularx}
\definecolor{iccvblue}{rgb}{0.21,0.49,0.74}
\usepackage[pagebackref,breaklinks,colorlinks,allcolors=iccvblue]{hyperref}

\usepackage[capitalize]{cleveref}
\crefname{section}{Sec.}{Secs.}
\Crefname{section}{Section}{Sections}
\Crefname{table}{Table}{Tables}
\crefname{table}{Tab.}{Tabs.}

\newcommand*{\affaddr}[1]{#1} 
\newcommand*{\affmark}[1][*]{\textsuperscript{#1}}

\renewcommand{\thefootnote}{\fnsymbol{footnote}}

\title{\ourmethod: Animated 3D Facial Avatar Generation from a Single Image}

\author{
Fei Yin\affmark[1,2], 
Mallikarjun B R\affmark[2],
Chun-Han Yao\affmark[2],
Rafał Mantiuk\affmark[1]\footnotemark[2],
Varun Jampani\affmark[2]\footnotemark[2],
\\
\affaddr{\affmark[1]University of Cambridge},
\affaddr{\affmark[2]Stability AI}
\\
}

\begin{document}
\maketitle

\renewcommand{\thefootnote}{\fnsymbol{footnote}}
\footnotetext[2]{Equal advising.}
\footnotetext[1]{
No biometric data was used to train, validate, or evaluate the model described in this work.}

\begin{abstract}
   \input{sections/0_abstract}
\end{abstract}

\input{sections/1_intro}
\input{sections/2_related_work}

\input{sections/3_method}

\input{sections/4_experiment}
\input{sections/5_conclusion}

{
\small
\bibliographystyle{ieeenat_fullname}
\bibliography{main}
}

\input{sections/X_suppl}

\end{document}

%% file: macros.tex
\newcommand{\ourmethod}{FaceCraft4D}

\newcommand{\refimg}{$I_R$ }
\newcommand{\coarseimg}{${I_i}$ }
\newcommand{\coarseimgs}{$\{ I_{i} \}$ }
\newcommand{\depthimgs}{$\{ D_{i} \}$ }
\newcommand{\refineimgs}{$\{ I_{i}^{*} \}$ }

\newcommand{\coinplaceholder}{COIN }

%% file: sections/0_abstract.tex
We present a novel framework for generating high-quality, animatable 4D avatar from a single image. While recent advances have shown promising results in 4D avatar creation, existing methods either require extensive multiview data or struggle with shape accuracy and identity consistency. To address these limitations, we propose a comprehensive system that leverages shape, image, and video priors to create full-view, animatable avatars. Our approach first obtains initial coarse shape through 3D-GAN inversion. Then, it enhances multiview textures using depth-guided warping signals for cross-view consistency with the help of the image diffusion model. To handle expression animation, we incorporate a video prior with synchronized driving signals across viewpoints. We further introduce a Consistent-Inconsistent training to effectively handle data inconsistencies during 4D reconstruction. Experimental results demonstrate that our method achieves superior quality compared to the prior art, while maintaining consistency across different viewpoints and expressions. 

%% file: sections/1_intro.tex
\section{Introduction}

\input{tables/tb_method}

4D Avatar generation aims to create animatable 3D head avatars, enabling the generation of consistent identities with custom viewpoints and expressions. High-quality 4D heads have significant applications in gaming, education, and film industries. 
However, existing 4D avatar building methods~\cite{teotia2024hq3davatar, qian2024gaussianavatars,ma20243dgaussianblendshapes,kirschstein2023nersemble} often require extensive inputs, such as multiview video with camera pose estimation, which is impractical for widespread use. Furthermore, these methods struggle to produce realistic results for extreme viewpoints and expression changes due to data scarcity, for instance the back view of the head.

The goal of this work is to create a 4D facial avatar from just a single image.
This task is challenging because a single image provides limited information, missing essential details like depth and multiple viewpoints that are crucial for constructing a complete 3D model.
A single image cannot provide information about the unseen parts of a person's head or capture dynamic aspects like facial expressions.
It is also impractical to train an end-to-end model due to the lack of multiview full-head animation datasets.
To address these challenges, the system needs to rely on prior knowledge, like shape and expression. 

\input{images/teaser}
Existing single-image avatar generation methods come with many shortcomings, which we summarize in Tab.~\ref{tb:method}. The works that rely on a single image of novel identity at test time, typically utilize large-scale, unstructured 2D images~\cite{an2023panohead} and videos~\cite{wei2024aniportrait,deng2024portrait4d} during training to account for multiview and temporal aspects. 
Due to the practical challenges in collecting multiview full-head animation datasets, most single-image methods, including recent work~\cite{taubner2024cap4d}, cannot reconstruct a complete 360-degree view of the head. While some approaches can generate full views~\cite{wu2024portrait3d,an2023panohead}, they typically lack the ability to model facial expressions.
Among methods that can animate, some methods rely solely on 2D representation~\cite{wei2024aniportrait}, which makes these methods multi-view inconsistent by design.
To overcome the above limitation, some methods rely on a hybrid representation that combines 2D and 3D~\cite{deng2024portrait4d,sun2023next3d}. However, such methods cannot handle extreme camera angles, as they rely on 2D components for high-resolution synthesis.
In contrast, we aim to build our model that has pure 3D representation, and, therefore, is multiview-consistent by design.

We propose \ourmethod{}, which takes a single image as input and creates a 3D Facial Avatar that can be animated and rendered in 360-degree view, as shown in Fig.~\ref{fig:teaser}.
To tackle such an under-constrained problem, we make use of 3 different types of priors: a shape prior, an image prior, and a video prior.
We use these priors to synthesize personalized multiview images of the given identity, which are then used to train an explicit 3D model of an avatar that can be animated (see Fig.~\ref{fig:pipeline}). We will refer to such a model as a 4D avatar. 

We start with a shape prior from pretrained 3D-GAN~\cite{an2023panohead}.
Since, 3D-GAN is trained with large-scale images that comprises of all the views of the head, it has information about the complete head shape. 
Through 3D-GAN inversion~\cite{roich2022pivotal}, we obtain personalized shape along with an approximate texture.
The resulting 3D representation lacks personalized high-quality multiview consistent texture and can not be used for creating novel expressions or animations (see Tab.~\ref{tb:method}). 
To improve the quality of the textures, we use of a 2D image prior. To ensure multi-view consistency and preservation of identity across the views, we employ epipolar constraints and cross-view mutual attention.
The resulting multi-view images can be used to build a high-quality static 3D head model, which, however, lacks control over expressions and articulations.
To animate the avatar, we utilize a 2D video prior~\cite{guo2024liveportrait}, which generates training data for facial animation across different viewpoints.

Finally, for real-time rendering performance, we train a 3D Gaussian representation~\cite{kerbl20233dgs} that is rigged to a FLAME parametric face model~\cite{FLAME:SiggraphAsia2017, qian2024gaussianavatars} using the multi-view videos obtained in the previous steps. To further improve the training consistency across the views and to avoid blur, we propose a COnsistent-INconsistent (COIN) training, which can capture cross-view inconsistencies in an MLP. 

In summary, the contributions of our work are as follows:
\begin{itemize}[leftmargin=*,noitemsep]
\item We present a novel framework for generating high-quality 4D head models from a single image of a face. We compare our framework with previous methods and summarize the differences in Tab.~\ref{tb:method}.
\item We introduce novel warping-based control generation signals to improve quality and consistency across different views and expressions. 
\item We introduce \coinplaceholder training, which enables a 3D Gaussian model to learn from inconsistent data while preserving high-quality, consistent features.
\end{itemize}

%% file: tables/tb_method.tex
\begin{table*}[t]
\begin{center}
\caption{
Comparison of methods for avatar generation and animation. 
The first two methods focus on general object generation, while the remaining methods are specifically designed for avatar applications. 
}
\label{tb:method}
\scalebox{0.75}{
\begin{tabular}{cccccccc}
\toprule
Methods & Goal & Dimension & Input & $360^\circ$ view & Animation & 3D Consistency & Quality \\ 
\midrule
LDM~\cite{rombach2022high}
& Image Generation & 2D & Single Image &  &  & - & High \\
DreamFusion~\cite{poole2022dreamfusion}, SV3D~\cite{voleti2024sv3d}
& Image-to-3D & 3D & Single Image & \Checkmark &  & Low & Low \\
\hline
LivePortrait~\cite{guo2024liveportrait}, AniPortrait~\cite{wei2024aniportrait}
& Video Generation & 2D & Single Image &  & \Checkmark & Low & High \\
PanoHead~\cite{an2023panohead}, SPI~\cite{yin20233d}
& 3D GAN Inversion & 3D & Single Image & \Checkmark &  & Medium & Medium \\
Portrait4D~\cite{deng2024portrait4d}, CAP4D~\cite{taubner2024cap4d}
& 3D Talking Head & 3D & Single Image &  & \Checkmark & Medium & High \\
\ourmethod & 4D Avatar Generation & 3D & Single Image & \Checkmark & \Checkmark & High & High \\
\hline
HQ3D~\cite{teotia2024hq3davatar}, GA~\cite{qian2024gaussianavatars}
& 4D Avatar Generation & 3D & Multi-view Videos & \Checkmark & \Checkmark & High & High \\
\bottomrule
\end{tabular}
}
\end{center}
\vspace{-15pt}
\end{table*}

%% file: images/teaser.tex
\begin{figure}[t]
\centering
\includegraphics[width=0.5\textwidth]{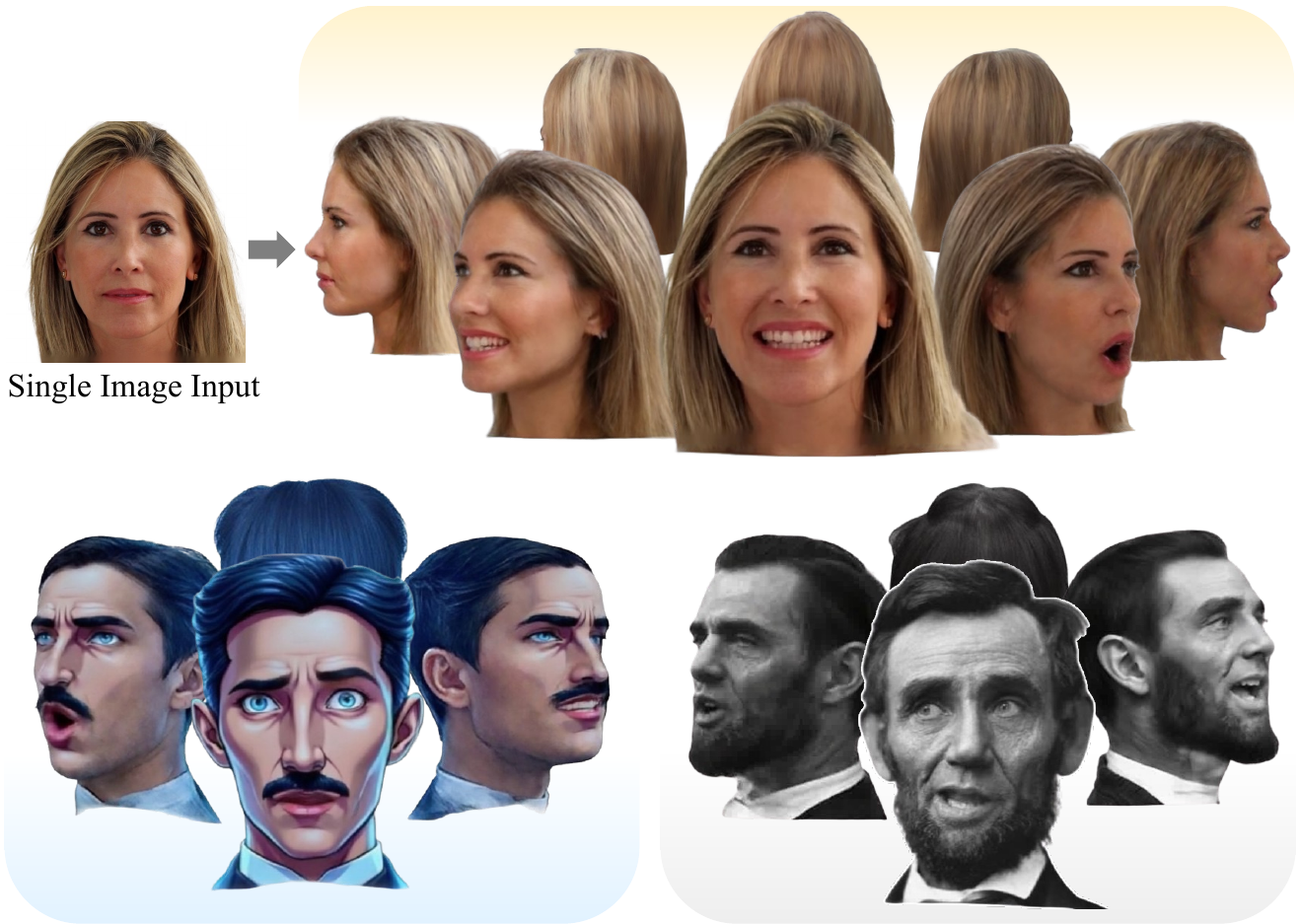}
\caption{Given a single image input, our method is capable of generating 4D avatars, producing photorealistic textures and consistent expressions across multiple views. Our approach also demonstrates robust performance on challenging inputs, including cartoon characters or old photographs.
}
\vspace{-0.5cm}
\label{fig:teaser}
\end{figure}

%% file: sections/2_related_work.tex
\section{Related Work}
\input{images/pipeline}

\noindent \textbf{3D Head generation.}
Numerous studies have explored the reconstruction of 3D head models. To obtain sufficient information, most approaches~\cite{grassal2022neural,zheng2023pointavatar,zheng2023pointavatar,qian2024gaussianavatars} utilize monocular video sequences, which provide multiple viewpoints of the subject's head. 
However, a significant limitation of these methods is their reliance on multiple viewpoints and lengthy video inputs. They also struggle to accurately reproduce less prominent features in videos, such as the back of the head.

Another approach is to train a 3D generator on a large head image dataset. EG3D~\cite{chan2022efficient} proposed a hybrid tri-plane representation that can quickly and efficiently render high-quality facial images. Panohead~\cite{an2023panohead} and Portrait3D~\cite{wu2024portrait3d} expanded on EG3D by increasing the number of planes, enabling the generation of 3D heads from limited angles to full 360-degree views. 
However, most methods in this category rely on hybrid representation having a 2D super-resolution network. This impacts their generalization to extreme camera poses and suffer identity inconsistency across views. 

Recently, the rise of text-to-image models has spurred the development of novel approaches for text-to-3D generation tasks.
One such approach is Score Distillation Sampling (SDS)~\cite{poole2022dreamfusion,raj2023dreambooth3d}, which leverages the score function to distill diffusion priors into 3D representations, enabling the generation of 3D content from text or image inputs.
Headsculpt~\cite{han2024headsculpt} incorporates facial keypoint information into the text-to-3D generation method, effectively constraining the facial layout.
However, the texture generated by these methods can be oversaturated, making it challenging to achieve photo-realistic head models. 

\vspace{1mm}
\noindent \textbf{Head animation.}
Head animation techniques can be broadly categorized into 2D-based and 3D-aware approaches, distinguished by their utilization of explicit camera models for rendering.
Recent methods~\cite{drobyshev2022megaportraits,hong2022depth,ren2021pirenderer,yin2022styleheat} have predominantly adopted warping-based strategies, applying deformation fields to intermediate appearance features to capture facial motion characteristics. 
While effective for frontal and moderate poses, these approaches struggle to maintain shape consistency during large pose variations due to their inherent lack of 3D modeling. 

To enable free-viewpoint rendering with stronger 3D consistency guarantees, researchers integrated explicit 3D representations into the head synthesis pipeline. Early approaches~\cite{khakhulin2022realistic,xu2020deep} relied on 3D morphable models for representing head shape and texture. Subsequent works~\cite{chu2024gpavatar,hong2022headnerf,li2023one,ye2024real3d,yu2023nofa} leveraged more sophisticated representations such as Neural Radiance Fields (NeRFs)~\cite{mildenhall2021nerf} to better capture complex structures like hair and accessories. Recent advances like GAGAvatar~\cite{chu2024generalizable}, which employs 3D Gaussian splatting~\cite{kerbl20233dgs}, have demonstrated impressive performance in both rendering quality and speed. However, these methods are typically limited in their ability to generate full 360-degree views.
Our approach enables true 360-degree view synthesis while maintaining real-time rendering capabilities and achieving superior animation quality.

%% file: images/pipeline.tex
\begin{figure*}[ht]
  \centering
  \includegraphics[width=1\textwidth]{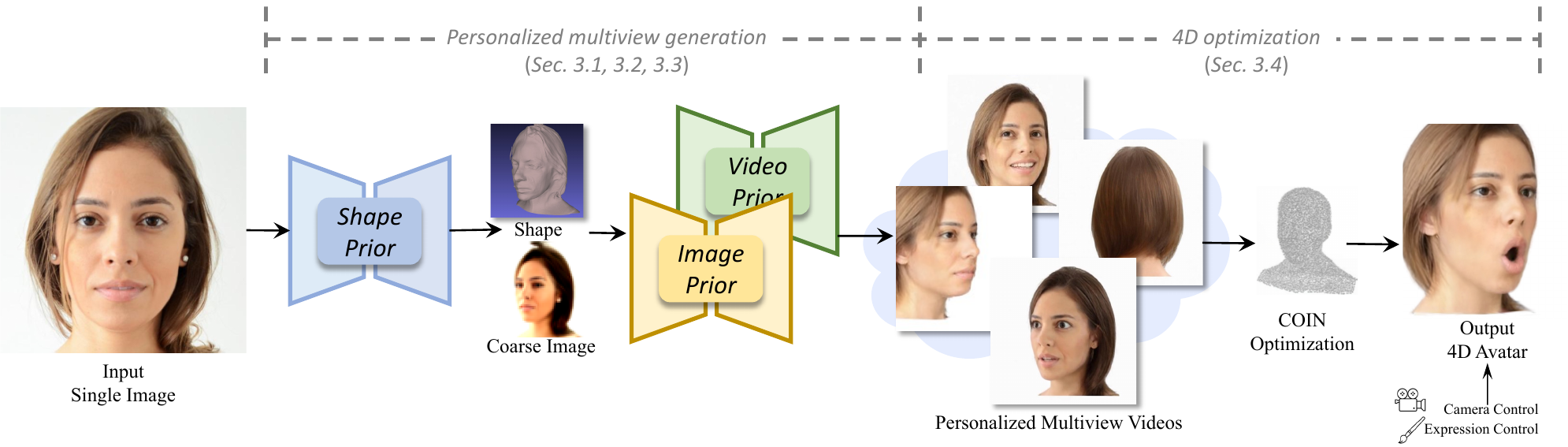}
  \caption{
  \textbf{Overview of \ourmethod}: Our approach begins with estimating the shape of a single input image using a shape Prior. 
  This shape guides the synthesis of personalized multiview images, providing $360^\circ$ views and varied expressions, with support from both 2D image and video priors.  
  Since the synthesized data often exhibit  inconsistency across views, we propose COIN optimization for robust 4D optimization.
  By fitting the model to the multiview data, we achieve our final 4D avatar. All faces in this manuscript come from the public FFHQ dataset~\cite{karras2019style}.
  } 
  \label{fig:pipeline}
  \vspace{-10pt}
\end{figure*}

%% file: sections/3_method.tex
\section{Method}

In this work, we propose a novel framework for generating high-fidelity 4D avatars from a single portrait image \refimg. Reconstructing a 3D dynamic model from a single 2D image is an inherently ill-posed problem, as critical 3D and temporal information is missing. To address this challenge, we leverage several 2D and 3D priors to compensate for the missing data.

The overall pipeline, shown in Fig.~\ref{fig:pipeline}, consists of two stages: personalized multiview generation and 4D representation optimization. In the Personalized Multiview Generation stage, we utilize shape priors to estimate the 3D shape (Sec.~\ref{sec:head_prior}), use image priors to enhance the texture (Sec.~\ref{sec:image_prior}), and video priors to synthesize high-quality, consistent multi-view videos with dynamic expressions (Sec.~\ref{sec:video_prior}). We then leverage this synthetically-generated personalized image set to optimize a robust 4D representation model (Sec.~\ref{sec:gaussian}).
The trained 4D model can be animated by the control parameters of a morphable model, enabling the generation of high-fidelity 4D avatars from a single portrait image. 

\subsection{Shape Prior}
\label{sec:head_prior}
The initialization is crucial for 3D generation. A good initialization should follow the approximate shape of the head shown in the input image, $I_R$. For that, we use the prior provided by a 3D-aware generator, PanoHead~\cite{an2023panohead,wu2024portrait3d}, which was trained on a large $360^{\circ}$ head dataset. 
We obtain a coarse 3D reconstruction for a given input image \refimg through GAN inversion.
Specifically, 
we first optimize the latent code $z$  within the generator's domain by minimizing a combination of pixel-wise $\mathcal{L}_2$ loss and image-level LPIPS loss~\cite{zhang2018perceptual}. 
Subsequently, following \cite{roich2022pivotal}, we keep the optimized latent code $z$ fixed while fine-tuning the 3D generator's parameters. 
During this fine-tuning stage, we maintain the same loss functions used in the latent optimization phase to ensure the rendered views closely match the input image. 
This process gives us an initial 3D head representation, though one that cannot be animated. 
The optimized generator is then used to synthesize a set of approximate multiview images \coarseimgs and their corresponding depth maps \depthimgs, where $i$ is the view index.

\subsection{Image Prior}
\label{sec:image_prior}
\input{images/image_prior}

\input{images/focal_length_comparison}
The textures synthesized from the shape prior may lack visual quality, and the identity of the individual may not be well preserved across the views. In particular, when rendering with different focal lengths, triplane-based methods~\cite{an2023panohead} exhibit substantial degradation in image quality, as illustrated in Fig.~\ref{fig:focal_length_comparison}. 
We enhance the generated multiview images using 2D diffusion models~\cite{rombach2022high}, which have demonstrated powerful generative capabilities.
Although direct image-to-image translation~\cite{meng2021sdedit} appears to be a straightforward solution, this approach poses significant risks: the denoising process may introduce cross-view inconsistencies and also modify the original semantic content. To address these risks, we propose two strategies that constrain the generation process, which we describe next and illustrate in Fig.~\ref{fig:image_prior}.

\vspace{1mm}
\noindent \textbf{Cross-view mutual attention}. 
Inspired by MasaCtrl~\cite{cao2023masactrl}, we introduce a cross-view attention mechanism to replace the standard self-attention during the generation process.
This modification allows us to inject information from the reference image into the novel view. 
Specifically, we first perform DDIM inversion~\cite{song2020ddim} to introduce noise to both the reference frame \refimg and the coarse image \coarseimg of the target novel view. This step preserves the overall face structure and contour.
Then, the diffusion model is used to gradually denoise two images to enhance the texture. In this process, we replace the key $K$ and value $V$ matrices (calculated in self-attention) of the novel view with those derived from the reference image. 
Different from \cite{cao2023masactrl}, we adopt a batch-processing strategy in which multiple images \coarseimg representing different viewpoints are treated as a single batch, with the reference image \refimg serving as the consistent source for all subsequent calculations. 
This not only streamlines the generation process but also ensures uniformity in feature injection across novel viewpoints.

The rationale behind this approach lies in the observation that, despite variations in viewpoint, the underlying textures (represented by the attention values) should maintain consistency across the views. The cross-view attention mechanism enables the transfer of information from the reference to the novel views, enhancing both the fidelity and coherence of the generated novel views. 

\vspace{1mm}
\noindent \textbf{Warping-based control generation}.
The consistency across the views is further enforced by epipolar constraints, which enforce shape relationships between novel views and the input view.
We first use cross-view mutual attention to generate the view of the back of the head. The generated back view along with the reference image are set as anchors.
Then, we propagate the texture information to novel views through depth-based warping.
Specifically, we lift the anchor image into 3D space and then project it to the adjacent view with depth guidance \depthimgs. 
During the projection, we form a mask to indicate the visibility of pixels to filter invisible areas using the rendered depth values.
The projected image is then used to guide the texture enhancement process.
The intermediate latent of the enhanced image in the diffusion model is blended with the latent of the projected image via a simple mask copy-pasting operation. 
By combining the latent, the resulting images benefit from both the improved consistency provided by the epipolar constraints and the enhanced quality achieved through the denoising process.
We repeat this process for various azimuth angles, creating a comprehensive set of viewpoints. 

With the aforementioned guidance and epipolar constraints, we can ensure coherence across the entire view space and generate high-fidelity view-consistent images \refineimgs for further reconstruction.

\subsection{Video Prior}
\label{sec:video_prior}
The view-consistent data created in the previous step has a static expression similar to that of reference image \refimg. This may not be enough to create diverse expressions as the images do not contain some of the expression-dependent information. For example, if the reference image has the mouth closed, the synthesized dataset would not have details inside the mouth region. To incorporate expression-dependent textures and shape that may not be available in the multiview image generation stage (Sec.~\ref{sec:image_prior}), we need to create multi-view facial expression data.
We use LivePortrait~\cite{guo2024liveportrait} as a video prior, which can help generate multi-view expressive videos. 
LivePortrait uses a single source image and a driving video to generate new videos where the source identity mimics the expressions and poses from the target video. 
Instead of directly copying the driving video's motion, one has the option of capturing the changes in relative expression but maintain the pose of the input image. 
Given that we have the reference and multiview-consistent enhanced images \refineimgs from the previous step, we provide these to LivePortrait as the source images together with driving videos with diverse poses and expressions (details in Sec.~\ref{subsec:implementation}), and obtain target videos.
Because the identity is consistent across the input views, the resulting videos also maintain the identity. Using identical driving videos ensures synchronized expressions. 

\input{sections/coin}

%% file: images/image_prior.tex
\begin{figure}[t]
\centering
\includegraphics[width=0.5\textwidth]{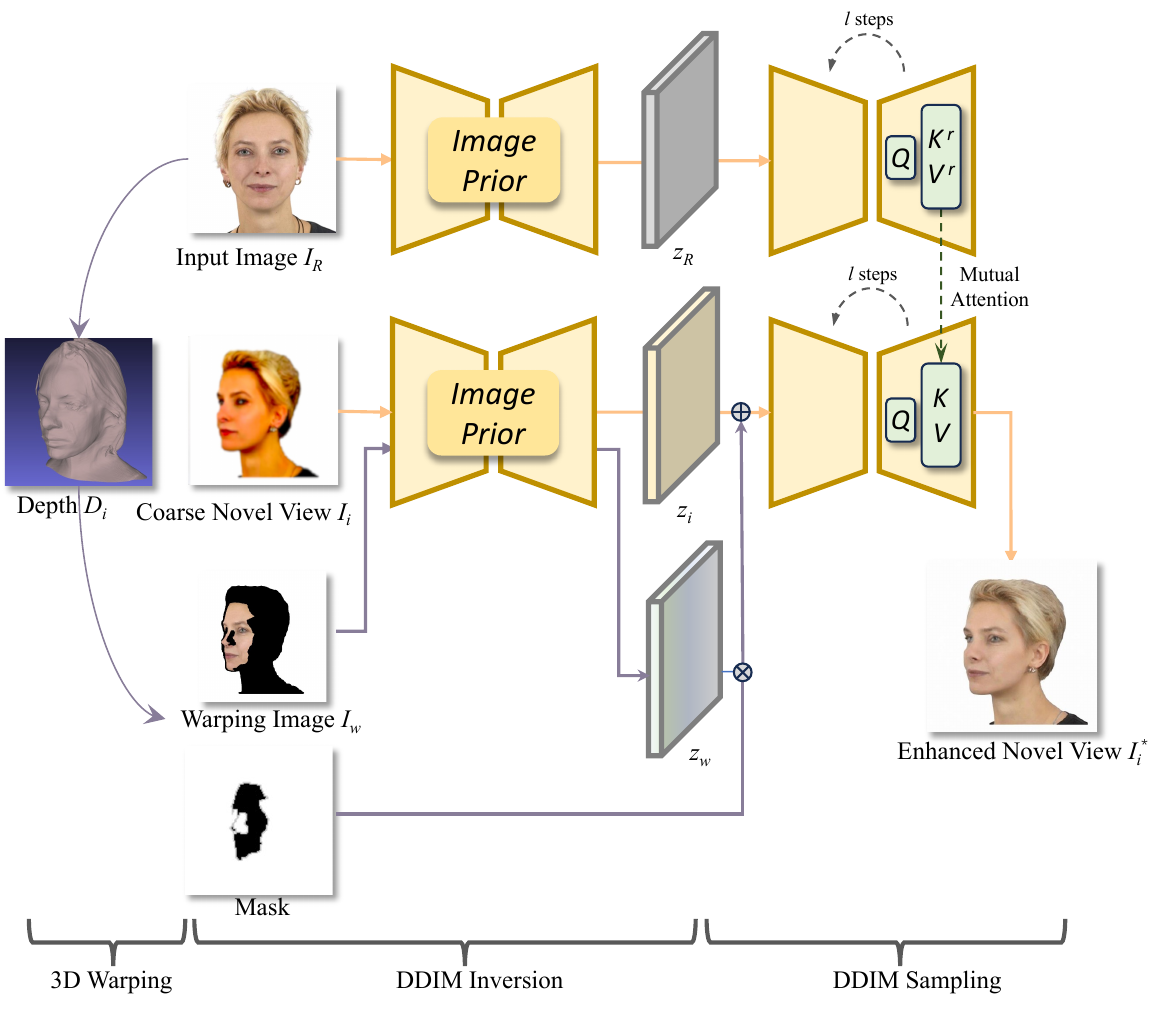}
\caption{\textbf{Image Prior}: We introduce a cross-view mutual attention mechanism and epipolar constraints to enhance consistency in generated novel views. Our approach aligns reference and target images, maintaining visual  coherence across viewpoints.}
\label{fig:image_prior}
\vspace{-10pt}
\end{figure}

%% file: images/focal_length_comparison.tex
\begin{figure}[t]
\centering
\includegraphics[width=0.5\textwidth]{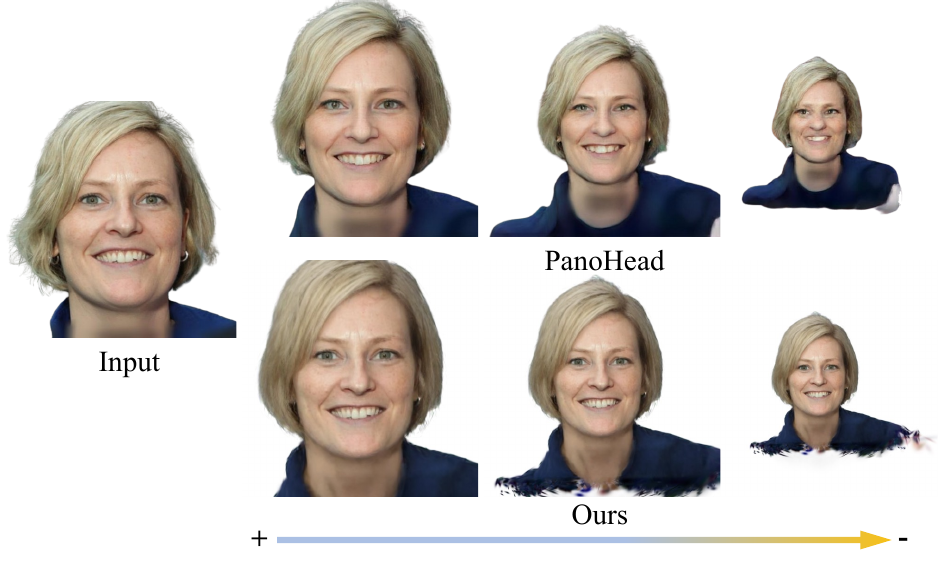}
\caption{
Triplane-based methods (\textit{e.g.} PanoHead) are sensitive to focal length (scale) and exhibit significant degradation of quality as focal length decreases. In contrast, our Gaussian-based methods maintain robust performance across varying focal lengths.
}
\label{fig:focal_length_comparison}
\vspace{-10pt}
\end{figure}

%% file: sections/coin.tex
\subsection{Robust 4D optimization with \coinplaceholder  training}
\label{sec:gaussian}

\input{images/gaussian}
\input{images/qualitative_reconstruction}
The synthesized multi-view videos have high-quality textures and good cross-view consistency, but small misalignment of features and colors is unavoidable. 
Learning 4D representation directly on such data leads to blurry textures for the rendered objects. 

Here, we address this problem by proposing a robust training paradigm.
At a high level, we learn a base representation that is common to all views, and a delta representation that is specific to each view during training.
This helps in cornering inconsistencies concerning each view to separate representation.
The base model will have a coarse but consistent representation learned, while the high-frequency details end up in the delta-space representation.
At test time, we choose one of the views as our reference view and use its delta representation as the common high-frequency details for all the views.

In specific, we achieve this by jointly training two 4D representations: a consistent GaussianAvatar~\cite{qian2024gaussianavatars}, which represents the 3D structure using Gaussians bound to a FLAME-tracked~\cite{FLAME:SiggraphAsia2017} mesh, and a NeRF-style MLP, tasked with encoding inconsistencies between the views. 
As shown in Fig.~\ref{fig:gaussian}, we design the losses that prioritize structure reconstruction in the consistent GaussianAvatar (LPIPS), and high-frequency detail reconstruction in the MLP with the view-dependent inconsistencies (L1 and SSIM). By isolating view-dependent variations in a separate model, this strategy prevents these inconsistencies from corrupting the base representation. The approach can be considered a form of robust regression, with the key difference that the view-dependent MLP ensures that inconsistencies (outliers) are spatially localized. This cannot be achieved with robust norms (\textit{e.g.}, L1), which do not maintain the information about the spatial position of outliers. 

We choose GaussianAvatar as our consistent representation because it lets us animate the avatar with controllable FLAME expression and pose parameters, providing a solid foundation for modeling consistent shape and appearance features.
The details of FLAME tracking and controlling are explained in Sec.~1.3 of the supplementary.
To model the view-dependent variations, we design a lightweight MLP-based architecture that introduces minimal computational overhead while effectively capturing view-specific details. The network encodes the offset in the color of each Gaussian (see Fig.~\ref{fig:gaussian}), which can be formulated as:
\begin{equation}
c_{\text{offset}}=\text{MLP}(e_{\text{view}},c,e_{g}; \theta),
\end{equation}
where $e_{\text{view}}\in \mathbb{R}^d$ is a learnable view-specific embedding that encodes view-representation,
$c$ represents the initial Gaussian color attributes from the GaussianAvatar, 
$e_{g}$ is a position embedding for each Gaussian, accounting for location-specific color variation ranges (e.g., eyes vs. mouth regions), and $\theta$ is a vector of the trainable parameters.
The final view-dependent color is computed as the sum of the base GaussianAvatar color and the offset: $c^{*}=c+c_{\text{offset}}$. 
The MLP consists of only two layers to maintain computational efficiency and preserve the real-time rendering capabilities of the Gaussian-based representation. 

During training, we render two types of images: $I^{\text{C}}_i$ from the consistent component alone, and $I^{\text{IC}}_i$ from the combination of both components. 
To ensure accurate reconstruction of high-frequency details, we supervise $I^{\text{IC}}_i$ using pixel-wise losses:
\begin{equation}
\mathcal{L}_{\text{pixel}}=\lambda_1 \mathcal{L}1(I^{\text{IC}}_i, I^*_i) + \lambda_{\text{SSIM}} \text{SSIM}(I^{\text{IC}}_i, I^*_i)
\end{equation}
To prevent the inconsistent component from dominating the representation and maintain the effectiveness of the consistent component, we introduce a structural supervision loss and a regularization term: 

\begin{equation}
\mathcal{L}_{\text{struc}}=\lambda_{\text{LPIPS}}\text{LPIPS}(I^{\text{C}}_i,I^*_i) 
\end{equation}

\begin{equation}
\mathcal{L}_{\text{reg}}=\lambda_{\text{offset}} \mathcal{L}1(c_{\text{offset}}, 0),
\end{equation}
where LPIPS loss ensures image-level supervision for the consistent component, and the offset regularization encourages minimal view-dependent modifications. 
The hyper-parameters are set to $\lambda_1=0.8$, $\lambda_\text{SSIM}=0.2$, $\lambda_\text{LPIPS}=0.05$ and $\lambda_\text{offset}=1$.

During inference, we input a single fixed view-embedding for the inconsistent component. Given that our reference image \refimg{} contains superior quality and identity information compared to the generated multiview images \refineimgs{}, we select its corresponding view embedding for inference. Novel expressions and poses can be easily generated by adjusting the FLAME parameters, leveraging the animation capabilities of our backbone model.

%% file: images/gaussian.tex
\begin{figure}[t]
\centering
\includegraphics[width=0.5\textwidth]{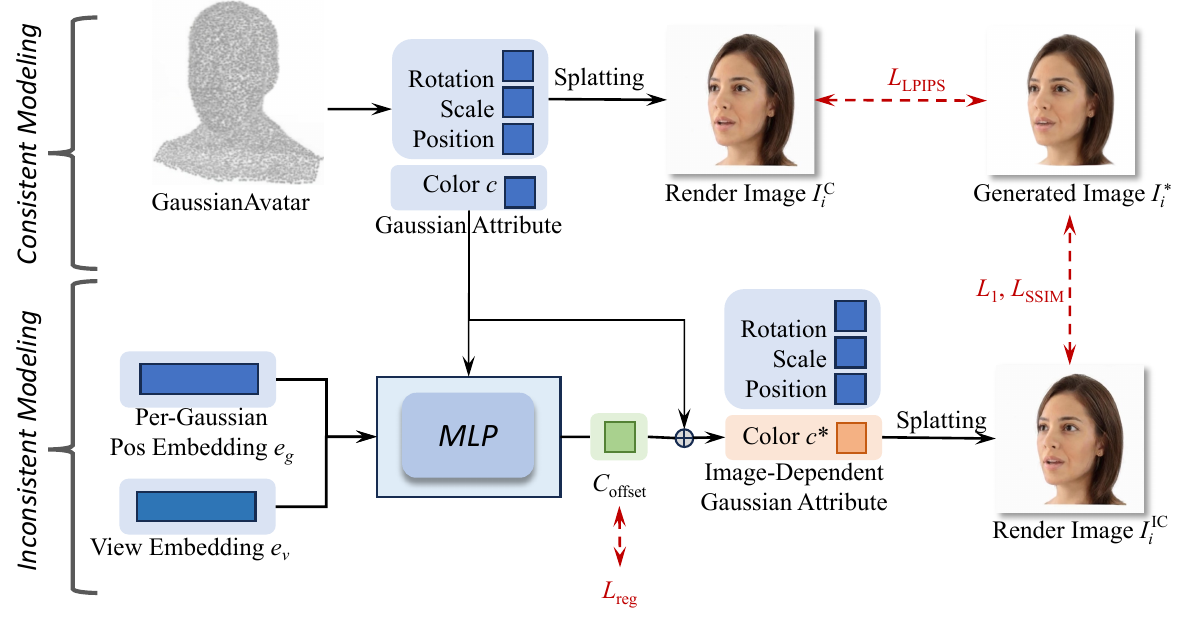}
\caption{\textbf{COIN-Training}: To address shape and color inconsistencies between the views, we train two 3D representations: a view-consistent base model (GaussianAvatar) and an MLP, encoding inconsistencies between the views. The MLP with inconsistencies lets us robustly reconstruct the high-quality 3D base model.}
\label{fig:gaussian}
\end{figure}

%% file: images/qualitative_reconstruction.tex
\begin{figure*}[ht]
  \centering
  \includegraphics[width=1\textwidth]{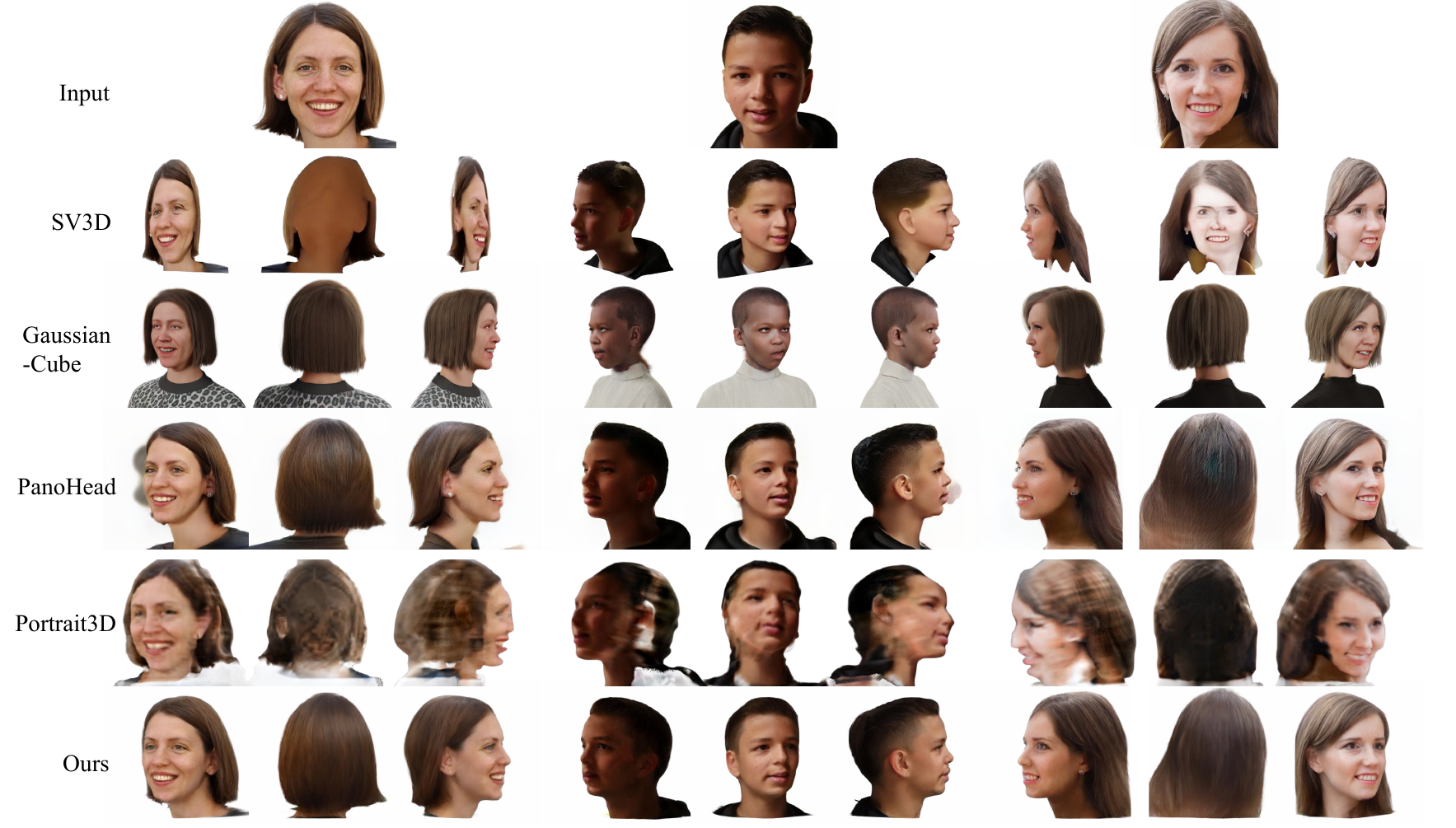}
  \caption{
  Qualitative comparison on static 3D head generation from a single image. 
  }
  \label{fig:qualitative_reconstruction}
  \vspace{-5pt}
\end{figure*}

%% file: sections/4_experiment.tex
\section{Experiments}

\input{images/qualitative_animation}
\noindent \textbf{Implementation details.}
\label{subsec:implementation}
During the multiview generation stage, we use PanoHead~\cite{an2023panohead} as our shape prior, Cosmicman~\cite{cosmicman} as the image prior, and LivePortrait~\cite{guo2024liveportrait} as the video prior model. We render 24 views along the horizon orbit for static reconstruction. For animation, we select \refimg and 11 frontal views from these generated images as source images and randomly sample 8 video clips from the NerSemble~\cite{kirschstein2023nersemble} dataset as driving videos. The source images and driving videos are then input to the video prior model to produce multiview video data.
In the 4D optimization stage, we first optimize the GaussianAvatar~\cite{qian2024gaussianavatars} model with the static data over 30K iterations. Next, we apply COIN-optimization techniques, finetuning the static Gaussian model over 90K iterations. The complete process takes approximately two hours to generate a single 4D avatar.

\vspace{1mm}
\noindent \textbf{Evaluation metrics.} 
In the absence of ground truth data, we use commonly adopted non-reference metrics to evaluate the rendered images in terms of image quality and identity similarity. For overall image quality, we use the FID~\cite{heusel2017fid} score. To assess multi-view consistency, we employ the CLIP-I~\cite{radford2021learning} score. For identity preservation, we calculate ID scores by measuring the cosine distance of face recognition features~\cite{deng2019arcface} between the novel views and the reference images.

\input{tables/tb_criterion}

\vspace{1mm}
\noindent \textbf{Datasets.} 
We conducted our analysis on a subset of facial images from FFHQ~\cite{karras2019style} dataset, a high-quality, in-the-wild facial dataset. For evaluation, we select 100 images with unoccluded faces to evaluate our model and baselines.

\subsection{Static 3D Generation Comparisons}

\vspace{1mm}
\noindent \textbf{Baselines.} 
We compare our method with existing image-to-3D approaches, including a general object-focused model, SV3D~\cite{voleti2024sv3d}, and three head-specific methods: the inference-based GaussianCube~\cite{zhang2024gaussiancube}, the 3D-GAN-based PanoHead~\cite{an2023panohead}, and the optimization-based Portrait3D~\cite{wu2024portrait3d}.

\vspace{1mm}
\noindent \textbf{Qualitative results.} 
Fig.~\ref{fig:qualitative_reconstruction} shows visual comparisons where our method achieves the highest visual fidelity and identity coherence compared to other approaches. Notably, it preserves details such as teeth and earrings in side views. In contrast, SV3D, which focuses on general objects, struggles to capture accurate head shape, often degenerating into a flat plane. GaussianCube tends to ``toonify" the input head due to the nature of its training data. While PanoHead produces appealing visual quality, it suffers from detail loss and floating artifacts during pose rotation. Portrait3D also falls short in quality, as the SDS~\cite{poole2022dreamfusion} optimization introduces variations that lead to blurring artifacts.

\vspace{1mm}
\noindent \textbf{Quantitative results.}
For each generated head model, we render images from five viewpoints: left-frontal, left, right-frontal, right, and frontal. All five rendered images, along with the reference image, are used to compute the CLIP-I, ID, and FID metrics.
As shown in Tab.~\ref{tb:reconstruction}, our method outperforms others in both ID and FID scores. This aligns with the observed visual fidelity, as our results appear more realistic and coherent. Our method achieves higher identity similarity and better detail preservation, even in side views.

\subsection{Animation Comparisons}

\vspace{1mm}
\noindent \textbf{Baselines.} We compare our method with one-shot video-based head reenactment approaches, including the 2D-based AniPortrait~\cite{wei2024aniportrait} and the 3D-based Portrait-4D~\cite{deng2024portrait4d}.
For AniPortrait, which uses keypoints as intermediate driving signals, we modify the Euler angles of the expression image to rotate the keypoints and generate novel view videos. In the case of Portrait-4D, novel view videos can be generated directly by editing the camera parameters.

\vspace{1mm}
\noindent \textbf{Qualitative results.}
As shown in Fig.~\ref{fig:qualitative_animation}, our method preserves expressions with greater accuracy. Portrait-4D lacks a back-of-head model, so when rotated to side views, the back appears empty. AniPortrait, as a 2D-based method, can achieve high texture quality. However, with large-angle rotations, AniPortrait struggles to preserve identity details and introduces noticeable artifacts in the hair and background. In contrast, our method maintains accurate shape and expression, even with camera movement.

\vspace{1mm}
\noindent \textbf{Quantitative results.} 
We test our data on unseen views (especially $60^\circ$–$180^\circ$), to highlight each method's performance under varying viewing angles, as handling extreme rotations is often challenging. 
The results are presented in Tab.~\ref{tb:animation}. Our method outperforms all baselines, demonstrating robustness across varied views. 
The performance of baselines degrades significantly as the viewing angle increases.

\input{images/ablation_reconstruction}
\input{images/ablation_animation}
\input{images/ablation_robustness}

\subsection{Ablation Study}
\vspace{1mm}
\noindent \textbf{Multiview image generation}.  
We conduct an ablation study on two multi-view generation modules, as shown in Tab.~\ref{tb:ablation_reconstruction} and Fig.~\ref{fig:ablation_reconstruction}. 
Incorporating the warping module allows us to preserve finer details, like tattoos, in novel views. 
While removing MA may lead to 
semantic inconsistencies, such as changes in gender. Adding mutual attention (MA) further improves fidelity and enhances identity similarity by injecting features from the reference images.

\vspace{1mm}
\noindent \textbf{Animation}.
Our 3D representation is based on GaussianAvatar~\cite{qian2024gaussianavatars}, which is FLAME-driven, enabling direct animation through FLAME parameters. To evaluate the base performance, we remove our video prior module. As shown in Fig.~\ref{fig:ablation_animation}, without the video prior, the final avatar fails to synthesize accurate expressions.
For consistency modeling, we compare our model with and without the COIN training. Without COIN, training directly on inconsistent generated images results in blurred textures. With COIN training, regions such as the teeth remain sharp, and finer details, like individual hair strands, are preserved. The quantitative results in Tab.~\ref{tb:ablation_animation} further validate the effectiveness of the COIN optimization across novel views.

\vspace{1mm}
\noindent \textbf{Processing times}. 
We analyzed the processing times of each stage.
Our method consists of three stages: a) 3D GAN inversion (2 mins); b) Multiview image \& video generation (10 mins), FLAME parameters extraction (10 mins); and c) COIN-training to overfit multiview videos (2 hours). The generation of a single avatar $\sim$2.5 hours, which is comparable to or faster than other optimization-based avatar reconstruction methods, as shown in Tab.~\ref{tb:time}. At inference, we achieve real time rendering at 156 FPS with a resolution of $512 \times 512$.

\vspace{1mm}
\noindent \textbf{Diverse inputs}. To validate robustness, we conduct experiments on diverse inputs, including cartoon, line drawing, and portraits with extreme camera poses. The results in Fig.~\ref{fig:ablation_robustness} demonstrate consistent performance across various domains.
Additional video results are available on the webpage in the supplementary materials.

\input{tables/tb_ablation_reconstruction}
\input{tables/tb_ablation_animation}
\input{tables/tb_time}

%% file: images/qualitative_animation.tex
\begin{figure*}[ht]
  \centering
  \includegraphics[width=1\textwidth]{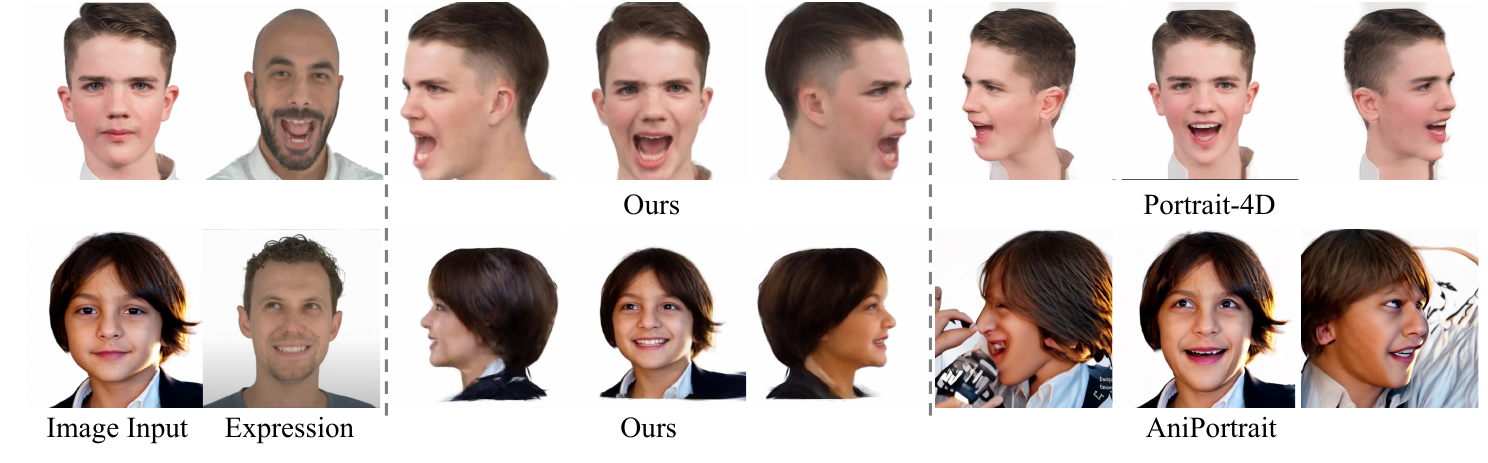}
  \vspace{-25pt}
  \caption{
  Qualitative comparison on animation of different views.
  }
  \label{fig:qualitative_animation}
  \vspace{-5pt}
\end{figure*}

%% file: tables/tb_criterion.tex
\begin{table}[t]
\centering
\caption{Quantitative comparison on static 3D head generation.}
\vspace{-10pt}
\label{tb:reconstruction}
\small
\setlength{\tabcolsep}{3.5mm}
\begin{tabular}{lccc}

\toprule
Method & CLIP-I $\uparrow$ & ID $\uparrow$ & FID $\downarrow$ \\ 
\midrule
GaussianCube~\cite{zhang2024gaussiancube} & 0.6830 & 0.4300 & 258.81 \\
PanoHead~\cite{an2023panohead} & \textbf{0.8233} & 0.4246 & 195.28 \\
SV3D~\cite{voleti2024sv3d} & 0.7656 & 0.4331 & 234.86 \\
Portrait3D~\cite{wu2024portrait3d} & 0.7066 & 0.3719 & 302.74 \\
Ours & 0.8053 & \textbf{0.5082} & \textbf{174.36} \\
\bottomrule
\end{tabular}
\vspace{-0.4cm}
\end{table}

\begin{table}[t]
\centering
\caption{Quantitative comparison on 3D head animation.}
\vspace{-10pt}
\resizebox{\linewidth}{!}{
\small
\setlength{\tabcolsep}{3.5mm}
\label{tb:animation}
\begin{tabular}{lccc}
\toprule
Method & CLIP-I $\uparrow$ & ID $\uparrow$ & FID $\downarrow$ \\ 
\midrule
AniPortrait~\cite{wei2024aniportrait} & 0.4653 & 0.4171 & 364.99 \\
Portrait-4D~\cite{deng2024portrait4d} & 0.5236 & 0.4592 & 248.36 \\
Ours & \textbf{0.5737} & \textbf{0.4602} & \textbf{201.76} \\
\bottomrule
\end{tabular}
}
\vspace{-15pt}
\end{table}

%% file: images/ablation_reconstruction.tex
\begin{figure}[t]
\centering
\includegraphics[width=0.5\textwidth]{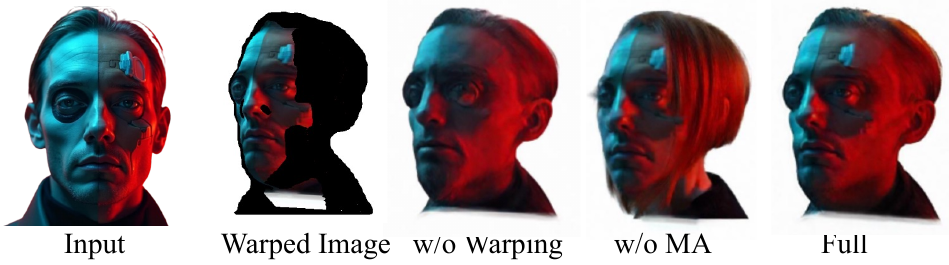}
\vspace{-20pt}
\caption{Ablation of multiview image generation modules.}
\label{fig:ablation_reconstruction}
\vspace{-10pt}
\end{figure}

%% file: images/ablation_animation.tex
\begin{figure}[t]
\centering
\includegraphics[width=0.5\textwidth]{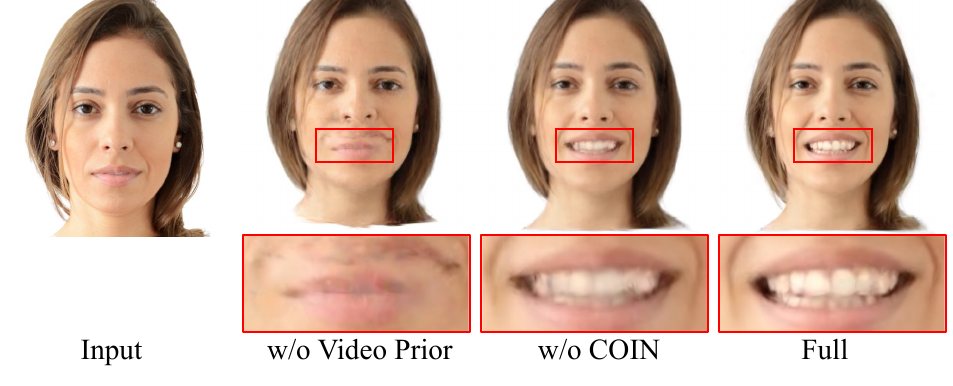}
\vspace{-7mm}
\caption{Ablation of animation modules.}
\label{fig:ablation_animation}
\vspace{-10pt}
\end{figure}

%% file: images/ablation_robustness.tex
\begin{figure}[t]
\centering
\includegraphics[width=0.5\textwidth]{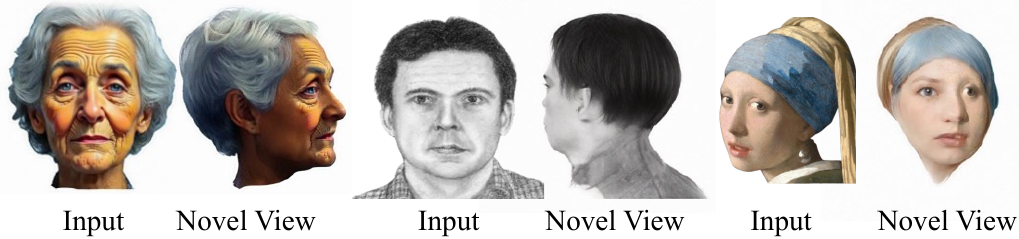}
\vspace{-20pt}
\caption{More results with diverse inputs.}
\label{fig:ablation_robustness}
\vspace{-10pt}
\end{figure}

%% file: tables/tb_ablation_reconstruction.tex
\begin{table}[t]
\centering
\caption{Ablation of multiview image generation modules. ``Warp" refers to the warping-based control generation (Sec.~\ref{sec:image_prior}) and MA to cross-view mutual attention (Sec.~\ref{sec:image_prior}).}
\vspace{-10pt}
\label{tb:ablation_reconstruction}
\small
\begin{tabularx}{\linewidth}{l>{\centering\arraybackslash}X>{\centering\arraybackslash}X>{\centering\arraybackslash}X>{\centering\arraybackslash}X}
\toprule
Method & CLIP-I $\uparrow$ & ID $\uparrow$ & FID \\ 
\midrule
w/o Warp w/o MA & 0.7328 & 0.4787 & 182.27 \\
w/o Warp & 0.7886 & 0.4915 & 171.08 \\
w/o MA & 0.8151 & 0.4951 & 172.43 \\
Full & \textbf{0.8162} & \textbf{0.4984} & \textbf{166.96} \\
\bottomrule
\vspace{-20pt}
\end{tabularx}
\end{table}

%% file: tables/tb_ablation_animation.tex
\begin{table}[t]
\centering
\caption{Ablation of animation modules.}
\vspace{-10pt}
\label{tb:ablation_animation}
\small
\begin{tabularx}{\linewidth}{l>{\centering\arraybackslash}X>{\centering\arraybackslash}X>{\centering\arraybackslash}X}
\toprule
Method & CLIP-I $\uparrow$ & ID $\uparrow$ & FID $\downarrow$ \\ 
\midrule
w/o COIN & 0.7688 & 0.4952 & 144.95 \\
Full & \textbf{0.7729} & \textbf{0.5010} & \textbf{142.80} \\
\bottomrule
\end{tabularx}
\vspace{-10pt}
\end{table}

%% file: tables/tb_time.tex
\begin{table}[t]
\centering
\caption{Time comparison with optimization-based avatar reconstruction methods.}
\label{tb:time}
\vspace{-3mm}
\scriptsize
\begin{tabular}{ccccc}
\toprule
Method & NerFace & IMAvatar & PointAvatar & HeadStudio \\
Time (hour) & 54h & 48h & 6h & 2h \\
\midrule
Method & MonoGA & GaussianAvatar & Hq3davatar & Ours \\
Time (hour) & 7h & 2h & 12h & 2.5h \\
\bottomrule 
\vspace{-0.6cm}
\end{tabular}
\end{table}

%% file: sections/5_conclusion.tex
\section{Conclusions}

In this paper, we present a novel method for generating high-quality 4D portraits from a single image. By leveraging multiple priors including 3D-GAN for shape initialization, image prior for texture enhancement, and video prior for animation, our method can generate animated avatars with detailed textures, excellent consistency, and plausible rendition of views missing in the input. The proposed depth-guided warping ensures cross-view consistency during the generation process, while our COIN-training strategy enables high-quality 4D reconstruction from potentially inconsistent multi-view data. Extensive experiments demonstrate that our method outperforms existing approaches in terms of shape accuracy, texture quality, and identity preservation across different viewpoints and expressions. Our framework makes high-quality 4D portrait generation more accessible by removing the requirement for multi-view data capture, paving the way for broader applications.

%% file: sections/X_suppl.tex
\clearpage
\setcounter{page}{1}
\maketitlesupplementary

\section{Implementation Details}
\subsection{Data Pre-processing}
Following \cite{an2023panohead}, we crop head regions for GAN inversion. Specifically, we use dlib~\cite{dlib09} to detect 68 facial keypoints. The keypoints are then aligned to ensure the face is centered in the image. 
To isolate the face, we apply matting to remove the background, replacing it with a white color. 
For SV3D~\cite{voleti2024sv3d}, which was trained on general objects, its training set typically centers objects in the image. To adapt to this domain, we add extra white padding around the face, ensuring better alignment with the model's original training conditions.

\subsection{Multiview Image Generation}
We set the DDIM inversion steps to $T=25$ and the image strength to 0.4, which corresponds to adding noise at timestep 10. Then, we use a diffusion model to denoise the added noise feature and apply cross-view mutual attention through the remaining steps.

In the warping-based control generation module, we employ a  copy-pasting operation to blend the warping features with the generation features. A mask is extracted during the warping process to identify regions in the novel view that correspond to reference textures in the original view. This mask is then downsampled to match the feature size. To handle boundary artifacts, we set the boundary values of the mask to 0.1. This adjustment mitigates the effects of disconnections or distortion artifacts commonly observed at the edges of warping features. By using a non-binary mask, we enable a smooth transition between the warping texture and the inpainted regions, effectively blending generative features with the warping values.

\subsection{Gaussian Training}

We choose GaussianAvatar~\cite{qian2024gaussianavatars} as our consistent Gaussian representation, which binds Gaussian features to a FLAME~\cite{FLAME:SiggraphAsia2017} template mesh. This approach allows animating the avatar with controllable FLAME expression and pose parameters.

We utilize the 2023 version of FLAME~\cite{FLAME:SiggraphAsia2017}, which includes revised eye regions. Additionally, following \cite{qian2024gaussianavatars}, we manually add 168 triangles to represent the teeth in the FLAME template mesh. The upper and lower teeth triangles are rigidly attached to the neck and jaw joints, respectively, improving the avatar's fidelity.

For FLAME tracking, we use the tracking code from \cite{qian2024gaussianavatars}. The optimization process includes per-frame parameters (translation, joint poses, and expression) and shared parameters (shape, vertex offset, and an albedo map). 
The optimization combines a keypoint loss, a color loss, and regularization terms. We optimize all the parameters on the first time step of the video sequence until convergence, then optimize per-frame parameters for 50 iterations for each following time step with the previous one as initialization. Afterward, we conduct global optimization for 30 epochs by randomly sampling time steps to fine-tune all parameters.
For more details, please refer to \cite{qian2024gaussianavatars}.

\begin{table}[t]
\centering
\caption{Quantitative comparison on static 3D head generation on MEAD~\cite{wang2020mead}.}
\resizebox{\linewidth}{!}{
\label{tb:supp1}
\small
\setlength{\tabcolsep}{3.5mm}
\begin{tabular}{lcccc}
\toprule
Method & PSNR $\uparrow$ & SSIM $\uparrow$ & LPIPS $\downarrow$ & ID $\uparrow$ \\ 
\midrule
Portrait3D~\cite{wu2024portrait3d} & 9.51 & 0.5622 & 0.5275 & 0.2646 \\
SV3D~\cite{voleti2024sv3d} & 13.54 & 0.6893 & 0.3174 &  0.7622 \\
PanoHead~\cite{an2023panohead} & 15.37 & 0.7342 & 0.2104 & 0.7981 \\
Ours & \textbf{15.74} & \textbf{0.7495} & \textbf{0.2010} & \textbf{0.8006} \\
\bottomrule
\end{tabular}
}
\end{table}

\section{Additional Experiments}

\subsection{Comparison with Controllable 3D GANs}
\begin{figure*}[t]
\centering
\includegraphics[width=1\textwidth]{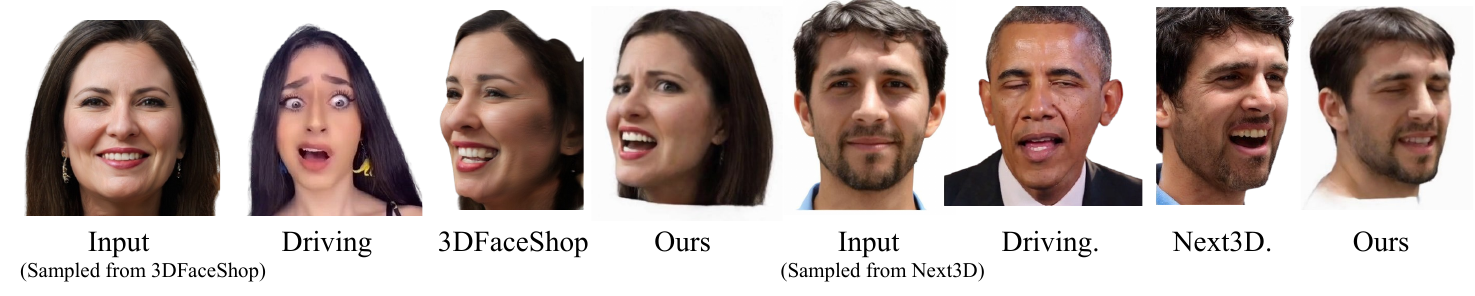}
\caption{Qualitative comparison with controllable 3D GANs.}
\label{fig:supp_3dgan}
\end{figure*}

We compare our method with controllable 3D GANs: 3DFaceShop~\cite{tang20233dfaceshop} and Next3D~\cite{sun2023next3d}. For a fair comparison, since controllable 3D GANs do not develop proper GAN inversion methods, we randomly sampled images from 3D GANs and used them as identity source images, then used other videos as expression signals. The qualitative results are shown in Fig.~\ref{fig:supp_3dgan}, demonstrating that our method achieves more accurate expressions from the driving images.

\subsection{Evaluation on Multiview Dataset}

\begin{figure*}[t]
\centering
\includegraphics[width=1\textwidth]{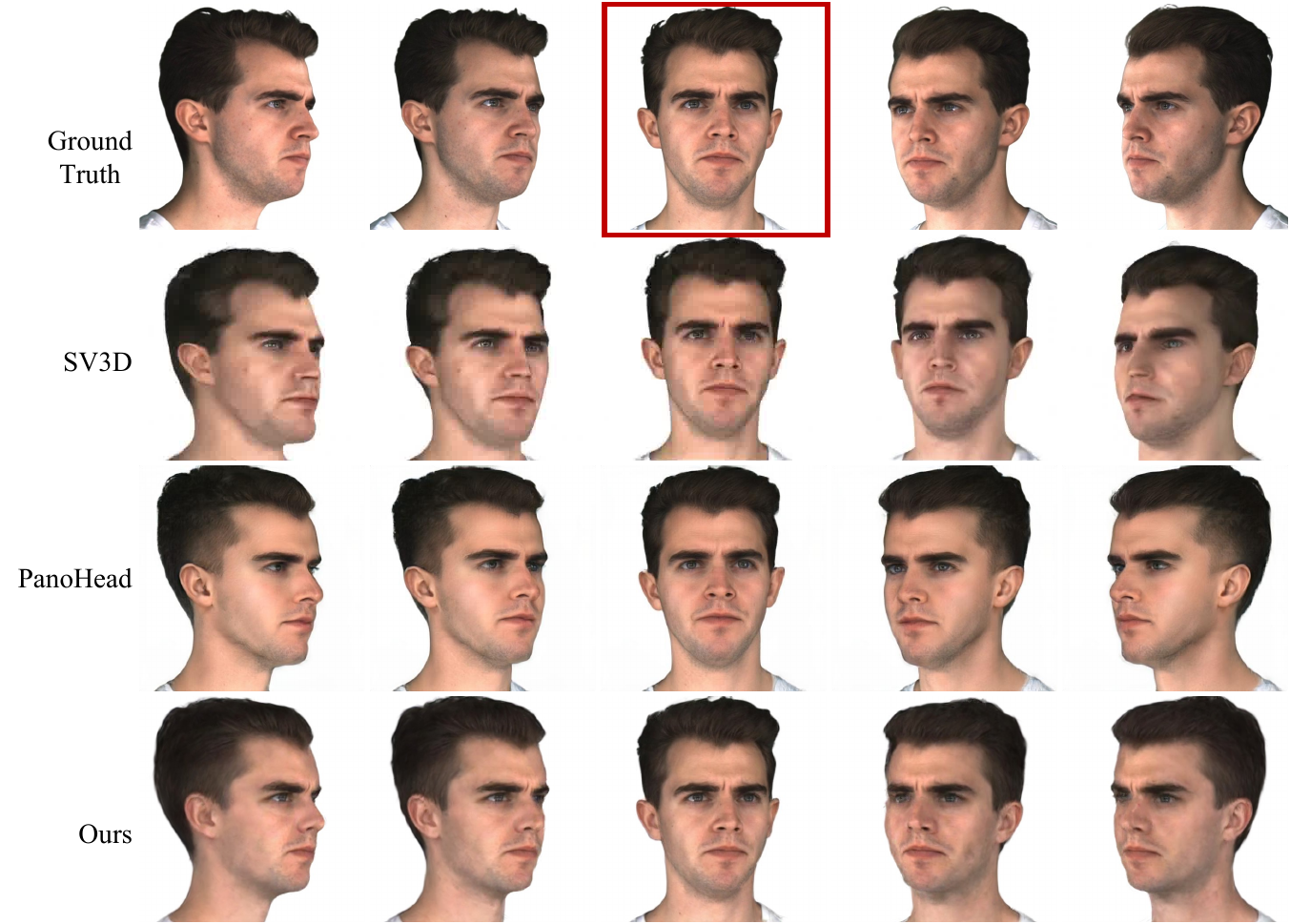}
\caption{Qualitative comparison on static 3D head generation from a single image on MEAD (1). Red box indicates the input image.}
\label{fig:supp1}
\end{figure*}

\begin{figure*}[t]
\centering
\includegraphics[width=1\textwidth]{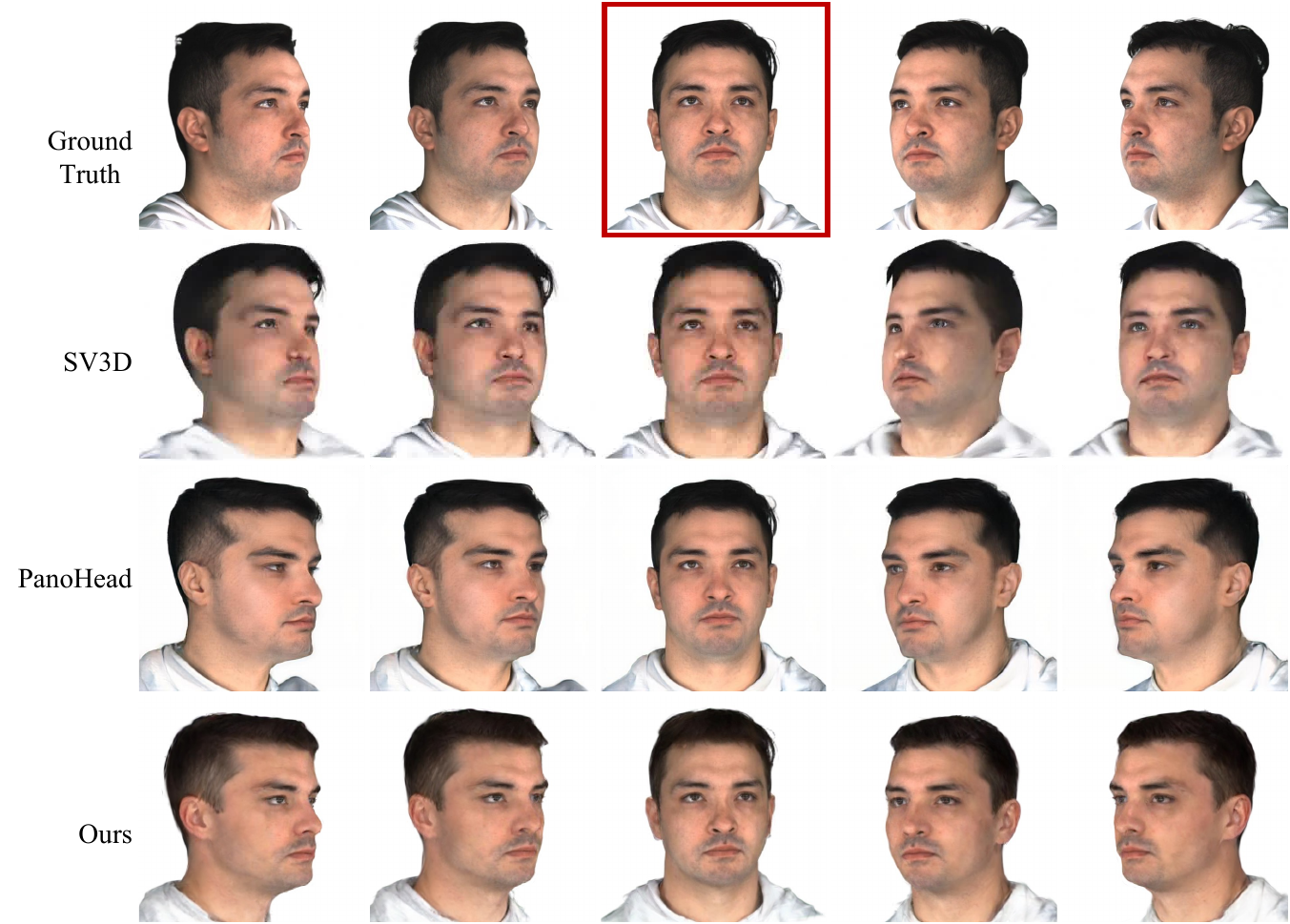}
\caption{Qualitative comparison on static 3D head generation from a single image on MEAD (2). Red box indicates the input image.}
\label{fig:supp2}
\end{figure*}

To get a comprehensive understanding of the performance of our method, we evaluate on MEAD~\cite{wang2020mead}, a multi-view dataset. The quantitative comparison between the reconstruction portraits and the ground truth is shown in Tab.~\ref{tb:supp1}. The qualitative results are shown in Fig.~\ref{fig:supp1} and Fig.~\ref{fig:supp2}. The results demonstrate that our method can generate a consistent shape and texture in novel views, which aligns with the conclusion of the manuscript.

\subsection{Video Results}
To further validate the robustness and real-time performance of our method, supplementary video recordings are provided. These videos demonstrate comparative experiments, accessible at:  
\url{https://github.com/yourname/project/videos}.

